# Thematic Dispersion in Arabic Applied Linguistics:

# A Bibliometric Analysis using Brookes' Measure

Ayman Eddakrouri, PhD

College of Humanities, Effat University,

Saudi Arabia

aeddakrouri@effatuniversity.edu.sa

Amani Ramadan, PhD

Faculty of Arts, Cairo University,

Egypt

amani@cu.edu.eg

**Date:** December 2025

---

## Abstract

This study applies Brookes' Measure of Categorical Dispersion ($\Delta$) to analyze the thematic structure of contemporary Arabic Applied Linguistics research. Using a comprehensive, real-world dataset of 1,564 publications from 2019 to 2025, classified into eight core sub-disciplines, we calculate a dispersion measure of $\Delta = 0.194$. This remarkably low value indicates extreme thematic dispersion, revealing that the field is characterized by pronounced heterogeneity rather than concentration. The analysis identifies Computational Linguistics as a dominant but non-hegemonic force, coexisting with robust research in Sociolinguistics, Language Teaching, and other subfields. This study clarifies the correct application of Brookes' original formula, demonstrates its utility for field characterization, and provides a replicable bibliometric methodology for assessing disciplinary structure across domains.

## 1. Introduction

The systematic analysis of scholarly literature through bibliometric methods has emerged as a crucial paradigm for understanding the intellectual organization and evolutionary dynamics of academic disciplines. As research output grows exponentially across all fields of inquiry, quantitative approaches provide essential tools for mapping knowledge structures, identifying research fronts, and assessing the maturity and focus of scholarly domains (Mongeon & Paul-Hus, 2016) (Zupic & Čater, 2015). Understanding whether research activity within a field is concentrated in specialized niches or dispersed across diverse subdomains has significant implications for curriculum development, funding allocation, journal scope definition, and identifying emerging areas of inquiry (Leydesdorff & Rafols, Indicators of the interdisciplinarity of journals: Diversity, centrality, and citations, 2011). Concentration suggests a mature, specialized field with well-defined paradigms, while dispersion may indicate either fragmentation or healthy interdisciplinary breadth.

Among the various bibliometric measures available, Brookes' Measure of Categorical Dispersion (Δ), introduced by B. C. Brookes in (Brookes, 1977), offers a particularly robust and parsimonious metric for quantifying this precise characteristic. Unlike more complex entropy-based measures or concentration ratios, Brookes' Measure provides an elegant, normalized metric that facilitates comparison across fields and time periods. However, despite its conceptual clarity and practical utility, the original formulation has occasionally been subject to misinterpretation and misapplication in subsequent literature, with some scholars introducing unnecessary complexities that obscure the measure's fundamental simplicity and mathematical elegance.

This study serves two primary and interconnected objectives. First, it employs this clarified metric in a substantive, empirical analysis of the thematic structure of Arabic Applied Linguistics research from 2019 to 2025. Second, it provides an authoritative clarification and step-by-step demonstration of the correct application of Brookes' original formula, addressing common points of confusion regarding ranking procedures and computational methods. By utilizing a substantial, real-world dataset comprising 1,564 publications across eight sub-disciplines, this research moves beyond hypothetical examples or small-scale demonstrations to deliver a comprehensive, data-driven assessment of the field's intellectual organization. The choice of Arabic Applied Linguistics as a case study is particularly apt both for its intrinsic interest as a rapidly evolving field confronting unique sociolinguistic complexities and because it constitutes the authors' primary area of scholarly expertise (Albirini, 2016) (Wahba, England, & Taha, 2018). This combination ensures the analysis is grounded in both robust methodology and deep domain knowledge, allowing us to move beyond hypothetical examples to deliver a comprehensive, data-driven assessment of the field's intellectual organization using a substantial, real-world dataset.

The findings of this analysis offer not only specific insights into the structure of Arabic Applied Linguistics but also provide a methodological template that can be directly applied to any scholarly domain, making a significant contribution to the toolkit of bibliometric practitioners and disciplinary analysts.

## 2. Brookes' Measure of Categorical Dispersion: Formula and Interpretation

### 2.1. The Original Formulation

Brookes' Measure ($\triangle$) represents a mathematically elegant solution to the problem of quantifying the degree to which items within a defined set are concentrated in a few categories versus being evenly distributed across all available categories. Developed within the field of information science, the measure finds particular application in bibliometrics for analyzing the distribution of publications across thematic subfields, journals, institutions, or other categorical frameworks.

The correct formula, as definitively presented by Brookes (1977), is:

$$\triangle = \frac{(M-1)}{(N-1)}$$

Where the variables are defined as follows:

a.   $N$: This represents the total number of thematic categories in the analysis. It is a simple count of non-empty categories being considered. In bibliometric applications, $N$ can represent the number of sub-disciplines, journals, or author affiliations under examination.

b.   $M$: This denotes the weighted mean rank of the frequency distribution. It serves as the core computational component that captures the shape of the distribution. $M$ is calculated using the formula:

$$M = \frac{\sum (f_i \times r_i)}{\sum f_i}$$

where:

o   $f_i$ = The absolute frequency (count) of publications in category i.

o   $r_i$ = The rank assigned to the category $i$.

c. **The Ranking Procedure (Crucial Clarification):** A common point of confusion involves the direction of ranking. **The category with the highest frequency ($f_i$) is assigned to the lowest rank, $r=1$.** This inverse relationship—where a higher frequency receives a lower rank number—is fundamental to the measure's logic. The second-highest frequency category receives $r = 2$, and so on, until all categories are ranked. In cases of tied frequencies, the average rank should be assigned to all tied categories (e.g., if two categories tie for the second-highest frequency, both receive a rank of 2.5).

## 2.2. Mathematical Properties and Interpretation

The measure produces a normalized ratio with well-defined theoretical bounds:

a. $\triangle = 1$ **(Theoretical Maximum Concentration):** This occurs when $M = N$. In practical terms, this extreme value implies that essentially all publications fall into a single top-ranked category ($r = 1$). The weighted mean rank $M$ approaches $N$, indicating minimal dispersion. This represents a hyper-specialized field where research activity is largely dominated by a single subdomain.

b. $\triangle = 0$ **(Theoretical Maximum Dispersion):** This occurs when $M = 1$. This scenario materializes when publications are perfectly evenly distributed across all $N$ categories. In such a case, the weighted mean rank approaches 1, signifying that no category dominates, and the field exhibits complete thematic breadth.

c. $0 < \triangle < 1$ **(The Empirical Spectrum):** All real-world distributions fall between these extremes. The value of $\triangle$ provides a precise location on the concentration-dispersion continuum:

- $\triangle > 0.5$: Indicates a tendency toward concentration. The field's research output is disproportionately focused on a subset of categories.

- $\triangle < 0.5$: Indicates a tendency toward dispersion. Research activity is spread relatively evenly across available categories.

- $\triangle \approx 0.5$: Suggests a balanced state between focus and diversity.

### 2.3. Advantages and Comparative Context

The primary advantage of Brookes' $\triangle$ lies in its **normalization** and **simplicity**. Because it is expressed as a ratio relative to ($N$-1), it allows for meaningful comparison across fields with different numbers of categories ($N$) and vastly different total publication volumes ($\sum f_i$). It is scale independent. This contrasts with raw measures, such as the Herfindahl-Hirschman Index (HHI) (Hirschman, September 1964) or the Gini coefficient (Gini, 1955) (Lorenz, 1905), which, while useful, require careful contextualization when $N$ varies. Brookes' $\triangle$ is specifically designed for categorical data and provides an immediately interpretable gauge of "concentration" within the specific context of the defined category system. Modern bibliometric studies continue to highlight the value of such normalized, comparative measures for mapping disciplinary structures (Leydesdorff & Bornmann, 2016).

---

## 3. Research Methodology

### 3.1. Theoretical Justification for the Time Frame (2019–2025)

The selection of the seven-year period from January 2019 to December 2025 is grounded in multiple methodological and substantive considerations that ensure the analysis captures a coherent and representative snapshot of the field's contemporary state.

a. **Capturing a Complete Research Cycle:** A seven-year window encompasses the duration of a full research funding cycle, a typical academic tenure-review period, and multiple publication cycles for major journals. This provides sufficient temporal depth to smooth out annual anomalies and reveal underlying structural patterns (Wahba, England, & Taha, 2018).

b. **Reflecting Methodological Maturation:** The period post-2018 coincides with the widespread consolidation of transformative methodologies in linguistics, including the mainstreaming of deep learning in Natural Language Processing (NLP) (Al-Kharashi, Alsubait, & Alahmadi, 2022) and the proliferation of large, accessible Arabic corpora.

c. **Incorporating Societal and Educational Shifts:** This time limit captures the global impact of the COVID-19 pandemic and its aftermath, which triggered profound, lasting changes in language pedagogy, assessment, and digital communication (Wahba, England, & Taha, 2018).

d. **Data Quality and Consistency:** Major bibliographic databases maintain more consistent indexing and coverage for recent publications, enhancing dataset reliability (Mongeon & Paul-Hus, 2016).

e. **Analytical Manageability and Relevance:** A seven-year corpus provides a substantial body of literature (n=1,564), large enough for robust statistical analysis while remaining current and relevant.

## 3.2. Data Collection, Cleaning, and Categorization Protocol

A systematic and reproducible data collection protocol was implemented:

a. **Source Selection:** The Scopus database was selected as the primary source due to its extensive coverage of peer-reviewed articles in Arabic Applied Linguistics (Abdullah, et al., 2025).

b. **Search Strategy:** A complex Boolean search query was developed to capture publications related to Arabic Applied Linguistics and its subfields, limited to 2019-2025.

c. **Data Cleaning:** The initial result set was exported and manually screened by the authors to remove false positives (e.g., papers mentioning "Arabic" only in references), non-linguistics papers, and duplicates. This process aligns with standard bibliometric cleaning procedures (Zupic & Čater, 2015).

d. **Categorization Framework:** Each publication in the cleaned dataset was assigned to a single, primary sub-discipline using a predefined, mutually exclusive classification scheme. This scheme was adapted from established taxonomies in applied linguistics (e.g., Schmitt & Celce-Murcia, 2002) but was pragmatically refined based on the observed content of the dataset. The eight final categories and their operational definitions were:

   i. **Computational Linguistics/NLP:** Research employing algorithms, models, or software for processing, analyzing, or generating Arabic language data.

   ii. **Sociolinguistics:** Research on the relationship between Arabic language use and social factors (e.g., identity, gender, policy, variation).

   iii. **Language Teaching:** Research focused on pedagogical methods, curriculum design, classroom practices, and teacher education for Arabic.

iv. **Discourse Analysis:** Research analyzing written or spoken Arabic texts beyond the sentence level, including critical discourse analysis, pragmatics, and conversation analysis.

v. **Second Language Acquisition (SLA):** Research on the cognitive and developmental processes of learning Arabic as an additional language.

vi. **Applied Linguistics (General):** Theoretical or overview papers that address the field broadly or do not fit neatly into other specific categories.

vii. **Corpus Linguistics:** Research primarily using corpus-based methods to describe patterns of Arabic usage, excluding purely computational/modeling papers.

viii. **Language Assessment:** Research on testing, evaluation, and measurement of Arabic language proficiency or skills.

e. **Reliability Assurance:** To ensure coding consistency, a stratified random sample of 15% of the publications (n≈235) was independently coded by a second rater with expertise in applied linguistics. Intercoder reliability was calculated using Cohen's Kappa (Cohen, 1960), yielding a score of κ = 0.87, which indicates "almost perfect" agreement according to standard benchmarks (Landis & Koch, 1977) (McHugh, 2012). Discrepancies were resolved through discussion to finalize the categorization.

## 3.3. Analytical Procedure

The analysis followed a strictly defined sequence of computational steps to apply Brookes' formula correctly:

a. **Frequency Tabulation:** Absolute counts ($f_i$) were calculated for each of the eight ($N$=8) categories.

b. **Rank Assignment:** Categories were sorted in descending order by frequency ($f_i$). The highest-frequency category was assigned rank ($r_i$) = 1, the next highest r=2, and so on down to $r$=8 for the lowest-frequency category.

c. **Calculation of Weighted Mean Rank (M):** For each category, the product ($f_i \times r_i$) was computed. The sum of these products $\sum (f_i \times r_i)$ was then divided by the total number of publications $\sum f_i$ to obtain $M$.

d. **Calculation of Brookes' $\triangle$:** The final measure was computed using the formula:

$$\triangle = (M - 1) / (N - 1) = (M - 1) / 7$$

e. **Interpretation and Contextualization:** The resulting $\triangle$ value was interpreted against the theoretical scale (0 to 1) and discussed in the context of the observed frequency distribution and the known characteristics of the field.

## 4. Results

The application of the methodology described above yielded the following empirical results.

### 4.1. Thematic Distribution

The foundational data for the analysis is presented in Table 1. It shows the absolute count of publications within each of the eight sub-disciplinary categories over the study period.

**Table 1: Thematic Distribution of Arabic Applied Linguistics Publications (2019-2025)**

| Sub-discipline (Category) | Frequency ($f_i$) | Proportion of Total (%) |
|---|---|---|
| Computational Linguistics | 767 | 49.0 |

| Sub-discipline (Category) | Frequency ($f_i$) | Proportion of Total (%) |
|---|---|---|
| Sociolinguistics | 264 | 16.9 |
| Language Teaching | 197 | 12.6 |
| Discourse Analysis | 127 | 8.1 |
| Second Language Acquisition | 76 | 4.9 |
| Corpus Linguistics | 53 | 3.4 |
| Applied Linguistics (General) | 46 | 2.9 |
| Language Assessment | 34 | 2.2 |
| **TOTAL** | $\sum f_i$ **= 1,564** | **100.0** |

### 4.2. Ranked Data and Computational Steps

To calculate Brookes' △, the data from Table 1 must be transformed by assigning ranks based on the descending order of frequency. This ranked dataset, along with the necessary intermediate calculations, is presented in Table 2.

**Table 2: Ranked Data and Calculation for Brookes' △**

| Rank ($r_i$) | Sub-discipline | Frequency ($f_i$) | Component ($f_i \times r_i$) |
|---|---|---|---|
| 1 | Computational Linguistics | 767 | 767 |
| 2 | Sociolinguistics | 264 | 528 |
| 3 | Language Teaching | 197 | 591 |
| 4 | Discourse Analysis | 127 | 508 |
| 5 | Second Language Acquisition | 76 | 380 |

| Rank ($r_i$) | Sub-discipline | Frequency ($f_i$) | Component ($f_i \times r_i$) |
|---|---|---|---|
| 6 | Corpus Linguistics | 53 | 318 |
| 7 | Applied Linguistics (General) | 46 | 322 |
| 8 | Language Assessment | 34 | 272 |
| **SUMS** | **$N = 8$** | **$\sum f_i = 1{,}564$** | **$\sum (f_i \times r_i) = 3{,}686$** |

### 4.3. Step-by-Step Calculation

Using the sums from Table 2, we proceed with the formal calculation.

a. **Calculate the Weighted Mean Rank ($M$):**

$$M = \frac{\sum (f_i \times r_i)}{\sum f_i} = \frac{3{,}686}{1{,}564}$$

$$M \approx 2.3568$$

b. **Calculate Brookes' Dispersion Measure ($\triangle$):**

$$\triangle = \frac{M - 1}{N - 1} = \frac{2.3568 - 1}{8 - 1} = \frac{1.3568}{7}$$

$$\triangle \approx 0.1938$$

**Result:** The Brookes' Measure for Arabic Applied Linguistics research from 2019 to 2025 is $\triangle \approx$ **0.194**.

## 5. Discussion

The calculated $\triangle$ value of **0.194** provides a powerful, single-number summary of the thematic structure of contemporary Arabic Applied Linguistics. This value is remarkably low, residing much closer to the theoretical minimum of 0 (maximum dispersion) than to the midpoint of 0.5 or the maximum of 1. This finding has profound implications for our understanding of the field's intellectual organization.

### 5.1. Interpretation of Extreme Dispersion ($\triangle \approx 0.19$)

A $\Delta$ value of 0.194 signifies a state of **extremely high thematic dispersion**. This indicates that publication activity is spread widely across the eight defined sub-disciplines rather than being concentrated in a narrow specialty. The field lacks a single, unifying core that attracts most research efforts. Instead, it functions as a broad, pluralistic arena where multiple, often methodologically distinct, research paradigms coexist and thrive simultaneously.

### 5.2. Structural Analysis of the Contributing Distribution

The dispersion Measure alone is informative, but its meaning is fully unlocked by examining the underlying distribution in Table 1.

a. **Dominant but Non-Hegemonic Node (Computational Linguistics):** With 767 publications (49% of total output), Computational Linguistics/NLP is the unequivocal leader. This reflects the global "computational turn" in language studies, and the specific challenges and opportunities presented by Arabic script and morphology. However, its 49% share, while large, is insufficient to create a concentrated field (which would require a share well over 70-80% to push $\triangle$ above 0.5). Its dominance is counterbalanced by substantial activity elsewhere.

b. **The Robust Middle Tier:** The significant contributions from **Sociolinguistics (16.9%), Language Teaching (12.6%), and Discourse Analysis (8.1%)** collectively account for over 37% of the literature. This demonstrates that traditional and socially oriented branches of applied linguistics remain extremely vital. They are not marginal pursuits but core components of the field's identity.

c. **The "Long Tail" and Its Effect:** The remaining four categories (SLA, Corpus Linguistics, General Applied Linguistics, Language Assessment) constitute a classic "long tail," making up about 13% of the output. In many concentration measures, such a tail is negligible. In Brookes' $\Delta$, however, the low frequencies but high ranks ($r=5$ to $r=8$) of these categories significantly increase the weighted mean rank M, thereby pulling the $\Delta$ value decisively downward toward 0. This mathematically formalizes the observation that the field sustains active, if smaller, research communities across a very wide range of specialisms.

## 6. Conclusion and Implications

### 6.1. Nature and Structure of Arabic Applied Linguistics

This high dispersion suggests that **Arabic Applied Linguistics is a "broad-tent" interdisciplinary space.** It accommodates, with apparently equal legitimacy:

a. **Techno-Scientific Inquiry:** Highly technical work in NLP and computational modeling.

b. **Social-Scientific Analysis:** Empirical and theoretical work in sociolinguistics and discourse analysis focusing on identity, power, and variation.

c. **Humanistic-Pedagogical Endeavor:** Research on teaching methods, curriculum, and the learner's mind.

This pluralism is likely not a sign of fragmentation but a necessary response to the **inherent complexity of the object of study.** The Arabic language context encompasses diglossia (Modern

Standard Arabic vs. dialects), significant dialectal diversity, a unique orthographic system, complex morphology, and politically charged sociolinguistic landscapes. No single methodological or theoretical approach can fully address this complexity. The observed dispersion, therefore, may represent a healthy, adaptive characteristic of a field tackling a multifaceted problem from multiple angles.

### 6.2. Practical and Strategic Implications of the Brookes' Measure

The value of Brookes' Δ extends beyond mere description.

a. **Benchmarking and Longitudinal Tracking:** △=0.194 establishes a precise baseline. Future studies can recalculate △ for 2026-2032 to determine if the field is converging (△ increasing due to, e.g., the overwhelming growth of one subfield) or diverging further (△ decreasing as new niches emerge). This allows for the quantitative tracking of disciplinary evolution.

b. **Comparative Field Analysis:** This value can be contrasted with △ calculated for other linguistic domains. One might hypothesize that **"Global English Applied Linguistics"** would show even lower dispersion (more subfields, more even distribution), while a niche like **"L2 Arabic Phonetic Acquisition"** would show a △ very close to 1 (high concentration). Such comparisons allow for macro-level mapping of disciplinary structures across academia (Leydesdorff & Bornmann, 2016).

c. **Informing Research Policy and Strategy:**

    a. For **journal editors**, the low △ suggests that a journal aiming for broad relevance in Arabic Applied Linguistics must actively solicit and publish work from across

this wide spectrum. A journal focusing solely on computational aspects, while potentially successful, would serve only one part of a highly dispersed community.

b. For **funding agencies**, it indicates that supporting a diverse portfolio of research across multiple subfields aligns with the field's intrinsic structure.

c. For **new scholars and graduate students**, it reveals a field with multiple, viable, and active research paths, not a single dominant paradigm to which one must conform.

### 6.3. Validation of Methodology

This analysis successfully demonstrates the practical application of Brookes' original formula. The clean, interpretable result ($\triangle=0.194$) emerges directly from a transparent process of frequency tabulation, inverse ranking, and a simple two-step calculation. This confirms the measure's utility as a succinct yet powerful descriptor of categorical distributions in bibliometrics. It underscores that correct application requires vigilance against overcomplication; the mathematical elegance of the original design is its greatest strength.

### 6.4. Limitations and Future Research

This study is limited by the coverage of the Scopus database, the inherent subjectivity in categorizing interdisciplinary research, and the selected timeframe. Future research should apply this methodology to other databases (e.g., Web of Science, Arabic-specific indices), extend the timeframe to track evolution, and conduct comparative studies with other world languages to contextualize the findings for Arabic Applied Linguistics.

## 7. Conclusion

This study has applied Brookes' Measure of Categorical Dispersion to a substantial, real-world bibliometric dataset in Arabic Applied Linguistics, yielding a dispersion measure of $\triangle \approx 0.194$. This result definitively characterizes the field, for the period 2019-2025, as exceptionally broad and thematically diverse. Computational Linguistics functions as a major attractor of research activity, but it exists within a vibrant ecosystem where sociolinguistic, pedagogical, and discourse-analytic approaches maintain a robust and substantial presence. The field's structure appears to be a functional adaptation to the complex realities of the Arabic language context.

More broadly, this research provides a definitive, step-by-step template for the correct application of Brookes' Measure, clarifying common points of confusion. By adhering to the original, parsimonious formulation— $\triangle = (M - 1)/(M - 1)$ with M derived from inversely ranked frequencies—researchers can obtain a valid, standardized, and highly interpretable metric suited for comparative analysis, longitudinal tracking, and strategic analysis of scholarly domains.

## 8. Acknowledgments

Funding: This work was supported by Effat University, Saudi Arabia, through its faculty research funding program. The funder had no role in the study design, data collection, analysis, interpretation, or writing of the manuscript.

---

## References

Abdullah, M. R., Talib, N. H., Zulkifli, M. F., Muhayudin, A. A., Abdallah, A. S., & Yaakob, K. M. (2025). A bibliometric analysis of scientific output in the field of Arabic Language:

Trends and research developments from 2005 to 2024. *International Journal of Modern Education, 7*(26). doi:https://doi.org/10.35631/IJMOE.726028

Albirini, A. (2016). *Modern Arabic Sociolinguistics: Diglossia, variation, codeswitching, attitudes and identity.* London: Routledge. doi:https://doi.org/10.4324/9781315683737

Al-Kharashi, I., Alsubait, T., & Alahmadi, T. (2022). Challenges in Arabic Natural Language Processing: A Survey. *International Journal of Artificial Intelligence and Applications, 13*(2), 15–28.

Brookes, B. C. (1977). A measure of categorical dispersion. *The Information Scientist, 11*(1), 11-17.

Cohen, J. A. (1960). A coefficient of agreement for nominal scales. *Educ. Psychol. Meas., 20*, 37-46.

Gini, C. (1955). Variabilità e mutabilità. In e. E. Pizetti and T.Salvemini, *Memorie di Metodologica Statistica.* Rome: Libreria Eredi Virgilio Veschi.

Hirschman, A. O. (September 1964). The Paternity of an Index. *American Economic Review*, 761-62.

Landis, J. R., & Koch, G. G. (1977). The measurement of observer agreement for categorical data. *Biometrics, 33*, 159-174.

Leydesdorff, L., & Bornmann, L. (2016). The operationalization of "fields" as WoS subject categories (WCs) in evaluative bibliometrics: The cases of "library and information science" and "science & technology studies". *Journal of the Association for Information Science and Technology, 67*(3), 707-714. doi:https://doi.org/10.1002/asi.23408

Leydesdorff, L., & Rafols, I. (2011). Indicators of the interdisciplinarity of journals: Diversity, centrality, and citations. *Journal of Informetrics, 5*(1), 87-100. doi:https://www.sciencedirect.com/science/article/abs/pii/S1751157710000854

Lorenz, M. O. (1905). Methods of measuring the concentration of wealth. *J. Amer. Statist. Assoc.*, 209–219.

McHugh, M. L. (2012). Interrater reliability: The kappa statistic. *22*(3), 276–282. doi:https://doi.org/10.11613/BM.2012.031

Mongeon, P., & Paul-Hus, A. (2016). The journal coverage of Web of Science and Scopus: a comparative analysis. *Scientometrics, 106*, 213–228. doi:https://link.springer.com/article/10.1007/s11192-015-1765-5

Schmitt, N., & Celce-Murcia, M. (2002). An overview of applied linguistics. In N. Schmitt, *An introduction to applied linguistics* (pp. 1-16). London: Arnold.

Wahba, K. M., England, L., & Taha, Z. A. (Eds.). (2018). *Handbook for Arabic Language Teaching Professionals in the 21st Century.* New York: Routledge. doi:https://doi.org/10.4324/9781315676111

Zupic, I., & Čater, T. (2015). Bibliometric methods in management and organization. *Organizational Research Methods, 18*(3), 429-472. doi:https://doi.org/10.1177/1094428114562629